\pdfoutput=1

\documentclass[11pt]{article}

\usepackage[]{emnlp2021}

\usepackage{times}
\usepackage{latexsym}
\usepackage{amsmath}

\usepackage{graphicx}
\usepackage{multirow}
\usepackage{hyperref}
\usepackage[T1]{fontenc}

\usepackage[utf8]{inputenc}

\usepackage{microtype}
\usepackage{enumitem}
\usepackage{xspace}

\newcommand{\dataset}[0]{BioNLI\xspace}
\title{\dataset: Generating a Biomedical NLI Dataset Using Lexico-semantic Constraints for Adversarial Examples}

\author{Mohaddeseh Bastan \\
  Stony Brook University \\
  \texttt{mbastan@cs.stonybrook.edu} \\\And
  Mihai Surdeanu \\
  University of Arizona\\
  \texttt{msurdeanu@email.arizona.edu} \\ \AND
   Niranjan Balasubramanian \\
  Stony Brook University\\
  \texttt{niranjan@cs.stonybrook.edu}
  }

\begin{document}
\maketitle
\begin{abstract}
Natural language inference (NLI) is critical for complex decision-making in biomedical domain. One key question, for example, is whether a given biomedical mechanism is supported by experimental evidence. This can be seen as an NLI problem but there are no directly usable datasets to address this. The main challenge is that manually creating informative negative examples for this task is difficult and expensive. We introduce a novel semi-supervised procedure that bootstraps an NLI dataset from existing biomedical dataset that pairs mechanisms with experimental evidence in abstracts. We generate a range of negative examples using nine strategies that manipulate the structure of the underlying mechanisms both with rules, e.g., flip the roles of the entities in the interaction, and, more importantly, as perturbations via logical constraints in a neuro-logical decoding system \cite{lu-etal-2021-neurologic}.

We use this procedure to create a novel dataset for NLI in the biomedical domain, called \dataset and benchmark two state-of-the-art biomedical classifiers. The best result we obtain is around mid 70s in F1, suggesting the difficulty of the task. Critically, the performance on the different classes of negative examples varies widely, from $97\%$ F1 on the simple role change negative examples, to barely better than chance on the negative examples generated using neuro-logic decoding.\footnote{Code and data is available at \url{https://github.com/StonyBrookNLP/BioNLI}}

\end{abstract}

\section{Introduction}
\label{intro}

%
%
Biomedical research has progressed at a tremendous pace, to the point where PubMed\footnote{\url{https://pubmed.ncbi.nlm.nih.gov}} has indexed well over 1M publications per year in the past eight years. Many of these publications include high-level mechanistic knowledge, e.g., protein-signaling pathways, which is critical for the understanding of many diseases \cite{ValenzuelaEscarcega2018LargescaleAR}, but which must be supported by lower-level experimental evidence to be trustworthy. 
Developing models that can understand and reason about such mechanisms is crucial for supporting effective access to the rich biomedical knowledge~\cite{bastan2022sume}. In particular, the current information deluge motivates the need for developing tools that can answer the question: ``Is a given mechanism supported by experimental evidence?''. This can be seen as a biomedical natural language inference (NLI) problem. Despite the prevalence of many biomedical NLP datasets~\cite{bionlp-2020-sigbiomed, bastan2022sume, krallinger2017biocreative}, there are no datasets that can be directly used to address this task.


\renewcommand{\arraystretch}{1.5}
\begin{table}
    \vspace{1em}
    \centering
    \small
    \begin{tabular}{| p{7.29cm}|}  \hline
         {\textbf{Premise:}The outflow of \textcolor{blue}{uracil} from the yeast Saccharomyces cerevisiae is known to be relatively fast in certain circumstances, to be retarded by \textcolor{violet}{proton} conductors and to occur in strains lacking a \textcolor{blue}{uracil} \textcolor{violet}{proton} symport. In the present work, it was shown that \textcolor{blue}{uracil} exit from washed yeast cells is an active process, creating a \textcolor{blue}{uracil} gradient of the order of -80 mV relative to the surrounding medium. Glucose accelerated \textcolor{blue}{uracil} exit, while retarding its entry. DNP or sodium azide each lowered the gradient to about -30 mV, simultaneously increasing the rate of \textcolor{blue}{uracil} entry. They also lowered cellular ATP content. Manipulation of the external ionic conditions governing delta mu H+ at the plasma membrane had no detectable effect on \textcolor{blue}{uracil} transport in yeast preparations thoroughly depleted of ATP.} \\ \hline
         {\textbf{Consistent Hypothesis:}It was concluded that \textcolor{blue}{<re> uracil <er>} exit is probably not driven by the s \textcolor{violet}{<el> proton <le>} gradient but may utilize ATP directly.} \\ \hline
         {\textbf{Adversarial Hypothesis:}It is concluded that \textcolor{violet}{<el> uracil <le>} exit from S. cerevisiae is an active process facilitated by a \textcolor{blue}{<re> proton <er>} gradient and ATP.} \\ \hline
         
    \end{tabular}
    \caption{Example of a premise/hypothesis pair in the \dataset dataset, as well as of an adversarial hypothesis that was automatically generated by an encoder-decoder network that manipulated the lexico-semantic constraints in the original hypothesis. Here the regulator entity is marked as \textcolor{blue}{<re> entity <er>}, and the regulated entity is marked as \textcolor{violet}{<el> entity <le>}. }
    \label{tab:example}
\end{table}

%
%
However, manually creating a biomedical NLI dataset that focuses on mechanistic information is challenging. Table~\ref{tab:example}, which contains an actual example from our proposed dataset, highlights several difficulties. First, understanding biomedical mechanisms and the necessary experimental evidence that supports (or does not support) them requires tremendous expertise and effort~\cite{kaushik2019learning}. For example, the premise shown is considerably larger than the average premise in other open-domain NLI datasets such as SNLI \cite{snli:emnlp2015}, and is packed with domain-specific information. Second, negative examples are seldom explicit in publications. Creating them manually risks introducing biases,  simplistic information, and systematic omissions~\cite{wu2021polyjuice}.

%
%
In this work, we introduce a novel semi-supervised procedure for the creation of biomedical NLI datasets that include mechanistic information. Our key contribution is automating the creation of negative examples that are informative without being simplistic. Intuitively, we achieve this by defining  lexico-semantic constraints based on the mechanism structures in the biomedical literature abstracts. Our dataset creation is as follows:
{\flushleft {\bf (1)}} We extract positive entailment examples consisting of a premise and hypothesis from abstracts of PubMed publications. We focus on abstracts that contain an explicit conclusion sentence, which describes a biomedical interaction between two entities (a regulator and a regulated protein or chemical). This yields premises that are considerably larger than premises in other open-domain NLI datasets: between 3 -- 15 sentences. 
{\flushleft {\bf (2)}} We generate a wide range of negative examples by manipulating the structure of the underlying mechanisms both with rules, e.g., flip the roles of the entities in the interaction, and, more importantly, by imposing the perturbed conditions as logical constraints in a neuro-logical decoding system \cite{lu-etal-2021-neurologic}. This battery of strategies produces a variety of negative examples, which range in difficulty, and, thus, provide an important framework for the evaluation of NLI methods. 

We employ this procedure to create a 
new dataset for natural language inference (NLI) in the biomedical domain, called \dataset. 
Table~\ref{tab:example} shows an actual example from \dataset.
The dataset contains 13489 positive entailment examples, and 26907 adversarial negative examples generated using nine different strategies.
An evaluation of a sample of these negative examples by human biomedical experts indicated that $86\%$ of these examples are indeed true negatives. 
We trained two state-of-the-art neural NLI classifiers on this dataset, and show that the overall F1 score     remains relatively low, in the mid 70s, which indicates that this NLI task remains to be solved. Critically, we observe that the performance on the different classes of negative examples varies widely, from $97\%$ accuracy on the simple negative examples that change the role of the entities in the hypothesis, to $55\%$ (i.e., barely better than chance) on the negative examples generated using neuro-logic decoding. Further, given how the dataset is constructed we can also test if models produce consistent decisions on all adversarial negatives associated with a mechanism, giving deeper insight into model behavior. Thus, in addition of its importance in the biomedical field, we hope that this dataset will serve as a benchmark to test models' language understanding abilities.

\section{Related Work}
\label{related}
Previous work on NLI in scientific domains include: medical question answering~\cite{abacha2016recognizing}, entailment based text exploration in health care~\cite{adler2012entailment}, entailment recognition in medical texts~\cite{abacha2015semantic}, textual inference in clinical trials~\cite{shivade2015textual}, NLI on medical history~\cite{romanov2018lessons}, and SciTail~\cite{khot2018scitail} which is created from multiple-choice science exams and web sentences.
These datasets either have modest sizes~\cite{abacha2015semantic}, target specific NLP problems such as coreference resolution or named entity extraction ~\cite{shivade2015textual}, and make use of experts in the domain to generate inconsistent data which is costly and labor-intensive. Additionally, they also focus on sentence-to-sentence entailment tasks, where both the premise and the hypothesis are no longer than one sentence. 
Most importantly, none of these are directly aimed at inference on mechanisms in biomedical literature.

Our work is also related to NLI tasks that go beyond sentence-level entailments. For example,~\citep{yin-etal-2021-docnli} include premises longer than a sentence, but only use three simple rule-based methods to create negative samples. \citep{yan-etal-2021-control, nie2019adversarial} use larger contexts as premises for the NLI task but only on general purpose domains like news, fiction, and Wiki. On the other hand, the \dataset dataset is an inference problem with large contexts as premises but in the biomedical domain which often requires handling more complex texts and domain knowledge.

There is also a growing body of research into exploring factual inconsistency in text generation models~\cite{maynez-etal-2020-faithfulness,zhu-etal-2021-faithfully,utama2022falsesum}. We take advantage of the known weakness of generation models for hallucination and also employ a constraint based neurological decoding from recently introduced decoding methods~\cite{lu-etal-2021-neurologic,lu2021neurologic,kumar2021controlled} to generate adversarial examples for \dataset dataset.

\section{\dataset Creation}
\label{task}
We model the task of understanding if a high-level mechanistic statement is supported by lower-level experimental evidence as 
natural language inference (NLI). The goal of NLI is to understand whether the given \textit{hypothesis} can be entailed from the \textit{premise} or not~\cite{dagan2005pascal}. This is typically modeled with three labels (entailed or not, plus a neutral class if the two texts are unrelated).
In our case, the premise contains the experimental evidence, while the hypothesis summarizes the higher-level mechanistic information. Both of these texts are extracted from abstracts of biomedical publications, where the beginning sentences (the {\em supporting set}) describe experimental evidence, and a {\em conclusion} sentence summarizes the mechanistic information that is entailed by these experiments. 


In this work, we introduce the \dataset dataset, an NLI dataset automatically created from a set of abstracts of PubMed open-access publications. We collected all the abstracts which contain a conclusion sentence with mechanistic information at the end of the abstract, and filter out the rest. Following previous work in mechanism generation~\cite{bastan2022sume}, we focus on conclusion sentences that discuss binary biochemical interactions between a regulator and a regulated entity (both of which are proteins or chemicals). We then generate negative examples by manipulating the structure of the conclusion sentences.


In the following subsections we describe in detail the generation of both positive and negative examples in \dataset.

\subsection{Identifying Abstracts with Mechanistic Information}

To identify abstracts that contain conclusion sentences with such binary biochemical interactions, we followed the same procedure and dataset~\footnote{This paper works at a higher level of abstraction, which contains causal semantic relations (or ``activations''), which are  always directed and not necessarily asymmetric.} as \cite{bastan2022sume} . That is, we used a series of patterns (e.g., finding words that start with \textit{conclud} all patterns are described in Appendix A) to identify conclusion sentences at the end of abstracts, and consider the previous ones as the supporting set. We analyzed the SuMe dataset and found that 91\% of the abstracts end with conclusion sentences, which indicates that the filtering heuristic is robust.


Further, we take advantage of the structured text in the biomedical domain, by focusing on abstracts that describe some mechanism between two biochemical entities.
One of the main entities is called \textit{regulator entity}  and is marked with \textit{<re> entity <re>} inside the text; the other main entity is called \textit{regulated entity} and is marked with \textit{<el> entity <le>} inside the text. We will use this structure to generate negative examples by modifying it.

\subsection{Positive Instances}
For positive examples, we simply use the original conclusion sentence from the abstract as the hypothesis and the supporting set as the premise. These sentences are likely to be accurate as they are written by domain experts, and also peer-reviewed by other scientists.

\subsection{Adversarial Instances}
\label{adver_data}
The key contribution of this paper is on the automatic creation of meaningful, yet difficult negative examples without the use of experts. We introduce multiple strategies for creating negative examples. We group these strategies into two groups: \textit{rule-based} and \textit{neural-based counterfactuals}, both of which are detailed below.
We show examples of these strategies in Table~\ref{tab:data_example}.

\subsubsection{Rule-Based Counterfactuals}
\label{rule_based}
This category consists of rule-based methods that convert a correct conclusion sentence (i.e., the hypothesis) into an instance that is not entailed by the given supporting set by perturbing parts of its semantic structure. Most of them are  used in general-domain factual consistency evaluating systems~\cite{kryscinski2019evaluating,zhu2020enhancing}:

{\flushleft \textbf{Swap Entity Names (SEN)}}: Swapping the entity names. This flips the roles of the entities in the interaction, i.e., the regulator becomes the regulated and vice versa, which contradicts the original evidence. 

{\flushleft \textbf{Swap Entity Positions (SEP)}}: In this perturbation we swap the positions of the two entities in text.  

{\flushleft \textbf{Swap Random Entity (SRE)}}: In this perturbation one of the main entities is randomly swapped with a different entity that occurs in the supporting set and has the same entity type. We use SciSpaCy~\cite{neumann-etal-2019-scispacy} and the built in  \textit{en\_ner\_bionlp13cg\_md} model to detect entity types. 

{\flushleft \textbf{Swap Random Entity with Out of text entity (SREO)}}: In this perturbation we swap one of the two entities in the interaction with a random entity from out of the context which was not available in the supporting set but has the same type as the main entity. Similarly, we use SciSpaCy with the same model to detect entity types. 

{\flushleft \textbf{Verb Negation (VNeg)}}: We randomly select one of the predicates in the original conclusion and change its polarity, e.g., from positive to negative or vice versa.

{\flushleft \textbf{Swap Numbers (SN)}}: If the conclusion contains a number, it is swapped with a different number, randomly chosen from the supporting set.

{\flushleft \textbf{Lexical Polarity Reversal (LPR)}}: We collected a list of terms describing mechanistic interactions (e.g., \textit{inhibition} and \textit{promotion}), and swapped them with their antonyms when encountered in the hypothesis.

\subsubsection{Neural-based Counterfactuals}
\label{neuralG}
The above methods are relatively simple perturbations, which might be easily detected by transformer-based classifiers. To counteract this potential limitation, we take advantage of transformer-based generation methods to create more complex and diverse set of negative examples. In this category we have two main approaches:

{\flushleft \textbf{Mechanism Generation (GEN)}}: We use a model pretrained on a mechanism generation task in the same context~\cite{bastan2022sume} to generate mechanism sentences (and the relation between main entities) for each abstract in our dataset. As our dataset overlaps with the one from \cite{bastan2022sume}, we implemented a 5-fold cross validation, and retrained the model with the corresponding training set in each fold. That is, for each split, we train with 4 folds and generate the output for the other fold. The generated texts which get a BLEURT score lower than $\lambda$, and predict the relation between two main entities incorrectly are selected as counterfactuals. Here, we set $\lambda=0.45$.

{\flushleft \textbf{Neurologic Decoding (GEN-ND)}}: Neurologic decoding is a decoding algorithm that enables neural language models to generate text while satisfying complex lexical constraints \cite{lu-etal-2021-neurologic}. We take advantage of this decoding method to impose different structure-aware constraints. For  generation,  we use the same model as the one described in GEN approach. For decoding this model,  we define the following constraints which result in generating negative examples. (we combine all three categories in our results table, naming the entire group \textit{GEN-ND}):

{\flushleft {\bf (1)} Neurologic Decoding with SEN Constraints (GEN-ND-SEN)}: We imposed as positive constraints (i.e., constraints that should be satisfied during decoding) that the two entities be present in the output, but we swapped their names. That is, the regulator and regulated entities are swapped; if both of them are satisfied in the generated text, the instance is used as a negative example. For example if the original conclusion has the following pattern: \\ 
\vspace{0.6em}
\scalebox{0.9}
{$ ...<re>entity1<er>...<el>entity2<le>$}

We add the following constraints to the neurological decoding:\\
\vspace{0.6em}
\scalebox{0.9}
{$[[<re> entity2 <er>],[<el> entity1 <le>]]$}\\
Compared with the general SEN introduced in section~\ref{rule_based}, by using these constraints we force the generation model to generate a natural yet negative and completely new sentence.

{\flushleft {\bf (2)} Neurologic Decoding with SRE Constraints (GEN-ND-SRE)}:
Similarly, we swapped one of the main entities in the positive constraints with a random entity from the supporting text. 
To make sure the generated sentence is not too similar to the original conclusion sentence, we used the generated sentence only if both constraints are satisfied and the semantic similarity between the text and the original text is smaller than $\delta$. To compute the semantic similarity we use BioLinkBERT~\cite{yasunaga2022linkbert}. We set $\delta$ to $0.9$.
    
{\flushleft {\bf (3)}   Neurologic Decoding with Negative Constraints (GEN-ND-NG)}: The third and last method we tried uses negative constraints, i.e., its lexical artifacts should {\em not} be present in the generated text.
In particular, the negative constraints we defined contained the original entities. By using the original entities as negative constraints the generated text receives a higher score if the main entities are not shown in their own roles (i.e., neither regulator nor regulated entities are not enclosed with specific markers).
\subsection{Dataset Statistics}
\setlength\tabcolsep{5.5pt}
\renewcommand{\arraystretch}{1.}
\begin{table}

    \centering

    \begin{tabular}{c|c|c|c|c|c}
        \multicolumn{2}{c|}{Dataset}&Train & Dev&Test&Sum  \\ \hline \hline
        \multicolumn{2}{c|}{+}&8489 &3000&2000&13489 \\ \hline
        \multirow{11}{*}{-} &SEN&2064&3000&2000&7064 \\
        & SEP&2022&3000&2000&7022 \\
        & SRE &81&20&15&116\\
        & SREO &1466 & 2584&1524&5574 \\
        &VNeg &837&1395&810&3042\\
        &SN&615&543&314&1472 \\
        & LPR&711&623&340&1674\\
        &GEN-ND&547&141&102&790\\
        &GEN&165&30&21&216\\ \cline{2-6}
        &Total &8508&11336&7126&26970\\
        & Unique& 8508&3000&2000&13508\\ \hline \hline
        \multicolumn{2}{c|}{Total} &16997&14336&9126&40459\\
        \multicolumn{2}{c|}{ Unique} &16997&3000&2000&21997\\

    \end{tabular}
    \caption{Dataset statistics of the larger distribution. Each instance is perturbed as many times as possible for the dev and test sets and once for the training set.}
    \label{tab:dataset_lrg}

\end{table}

The resulting dataset is summarized in two tables. Table~\ref{tab:dataset_lrg} shows the maximum number of possible perturbations on each instance. For example, all instances can be perturbed with SEP and SEN approaches, while only the ones that have a number in both conclusion and supporting set can be perturbed with the SN approach.


\begin{table}
    \centering

    \begin{tabular}{c|c|c|c}
        \multicolumn{2}{c|}{Dataset}&Train & Dev  \\ \hline \hline
        \multicolumn{2}{c|}{+}&2790 &2453 \\ \hline
        \multirow{11}{*}{-} &SEN&413&2453 \\
        & SEP&452&2453 \\
        & SRE &80&22\\
        & SREO &335 & 2058 \\
        &VNeg &159&1214\\
        &SN&256&625 \\
        & LPR&311&728\\
        &GEN-ND&586&84\\
        &GEN&162&33\\ \cline{2-4}
        &Total &2754&9670\\
        &Total Unique& 2754&2453\\ \hline \hline
        \multicolumn{2}{c|}{Total} &5544&12123\\
        \multicolumn{2}{c|}{Total Unique} &5544&2453\\

    \end{tabular}
    \caption{Dataset statistics of the balanced distribution. We sampled over perturbed classes to create a balance dataset so that no rule-based category have more than 500 instances in the train set. Test set is same as Table~\ref{tab:dataset_lrg}}
    \label{tab:dataset_blnc}
\end{table}

 \begin{figure}
     \centering
     \includegraphics[width=0.48\textwidth]{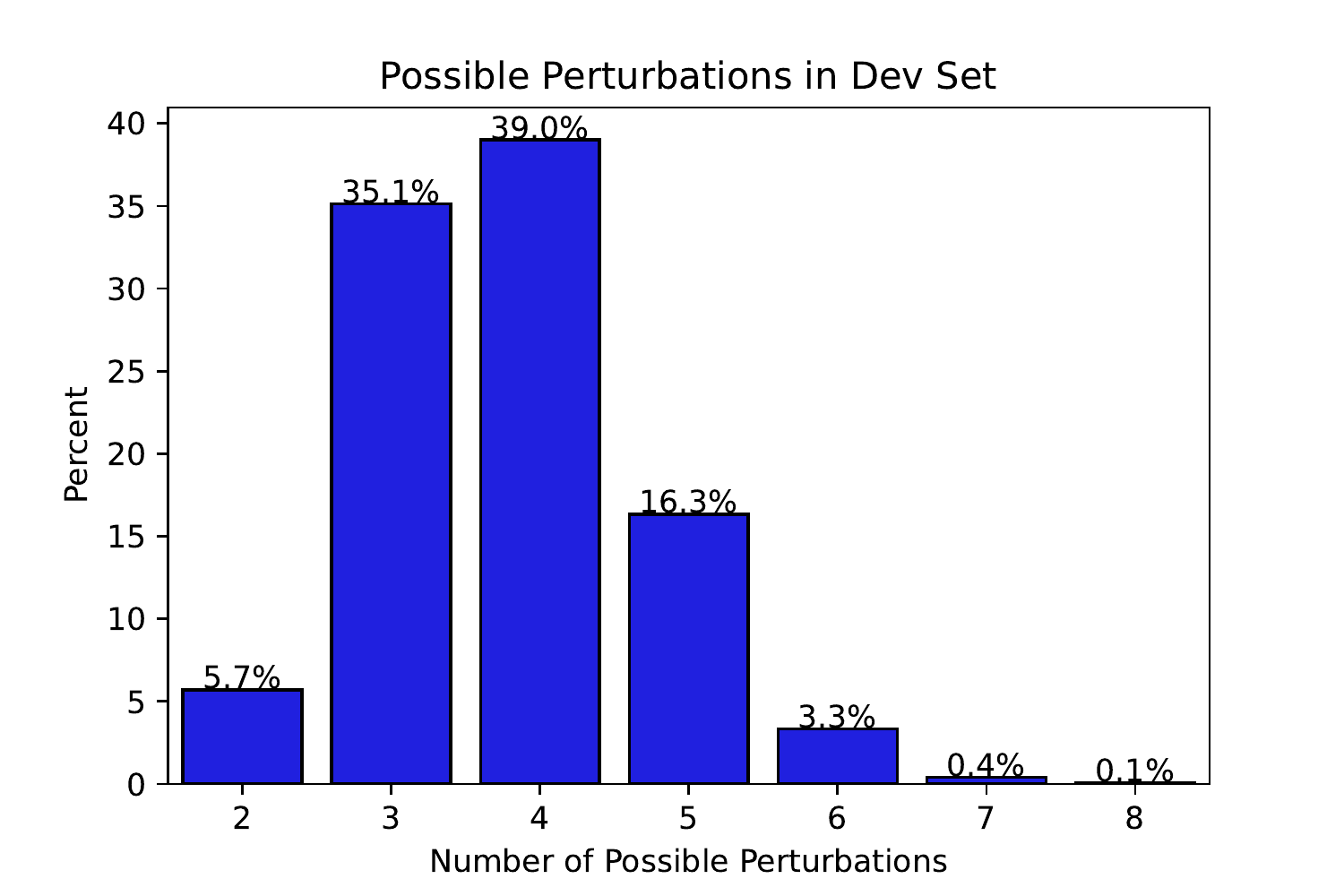}
     \caption{Distribution of possible perturbations over the dev set}
     \label{fig:dist_pert}
 \end{figure}

The distribution of the possible perturbations over the dev set is shown in Figure~\ref{fig:dist_pert}. As we see in this figure, all sentences can get at least two perturbation SEN and SEP. $5.7\%$ of the instances can not get any other perturbations based on their structure. $35.1\%$ of the data get 3 different adversarial examples. Most of the data, which is about $39\%$ of them, can be perturbed with 4 different approaches. The lowest category is 8 perturbations which are only about $0.1\%$ of the data. We don't have any instance which can be perturbed with all 9 possible methods explained in section~\ref{adver_data}, which shows the diversity and variety on the adversarial instance generation approaches.

Note that while our goal is to produce a dataset with as many high quality examples as possible for each category, downstream applications can adjust the distribution of the training categories to be uniform or biased towards specific categories as needed based on their requirements. To study the impact of a balanced distribution of adversarial examples, we sampled the positive and negative classes to produce a balanced dataset. Table~\ref{tab:dataset_blnc} shows the distribution of the adversarial categories in this balanced dataset. We evaluate with the original collection as well as with this balanced dataset (see Table~\ref{tab:per_class_perf}).\\
Table~\ref{tab:data_example} shows a set of rule-based and neural-based adversarial examples. The main entities are enclosed with specific markers and are swapped with different methods. We also see a completely new and negative generated text based on the supporting set with the generation approach (GEN-SEN).

\subsection{Quality Control}
To ensure the quality of the collected dataset, we asked two experts (graduate students in the biomedical domain) to inspect 50 randomly selected adversarial examples generated by the neural-based counterfactual methods (section~\ref{neuralG}). We sampled 25 examples each from GEN and GEN-ND methods. Given the abstract and the generated sentence, the experts assessed whether the generated sentence is indeed inconsistent with the given supporting set, meaning that the sentence cannot be concluded given the supporting set. The expert analysis shows that the neural-based counterfactuals are of high quality. They find that $88\%$ of adversarial examples from the GEN method and $84\%$ from the GEN-ND method are correct negative examples, averaging to $86\%$ overall.

\renewcommand{\arraystretch}{1.5}
\begin{table*}[ht!]
    \centering
    \small
    \begin{tabular}{c | p{0.45\linewidth}| p{0.4\linewidth}}  
         \multicolumn{3}{p{0.96\linewidth}}{\textbf{Abstract:}We investigated whether intracellular pH (pH(i)) is a causal mediator in abscisic acid (ABA)-induced gene expression. We measured the change in pH(i) by a "null-point" method during stimulation of barley (Hordeum vulgare cv Himalaya) aleurone protoplasts with ABA and found that ABA induces an increase in pH(i) from 7.11 to 7.30 within 45 min after stimulation. This increase is inhibited by plasma membrane H(+)-ATPase inhibitors, which induce a decrease in pH(i), both in the presence and absence of ABA. This ABA-induced pH(i) increase precedes the expression of RAB-16 mRNA, as measured by northern analysis. ABA-induced pH(i) changes can be bypassed or clamped by addition of either the weak acids 5,5-dimethyl-2,4-oxazolidinedione and propionic acid, which decrease the pH(i), or the weak bases methylamine and ammonia, which increase the pH(i). Artificial pH(i) increases or decreases induced by weak bases or weak acids, respectively, do not induce RAB-16 mRNA expression. Clamping of the pH(i) at a high value with methylamine or ammonia treatment affected the ABA-induced increase of RAB-16 mRNA only slightly. However, inhibition of the ABA-induced pH(i) increase with weak acid or proton pump inhibitor treatments strongly inhibited the ABA-induced RAB-16 mRNA expression.} \\ \hline
         \multicolumn{3}{p{0.96\linewidth}}{\textbf{Conclusion:}We conclude that, although the \textcolor{violet}{<el> ABA <le>}-induced the \textcolor{blue}{<re> pH <er>}(i) increase is correlated with and even precedes the induction of RAB-16 mRNA expression and is an essential component of the transduction pathway leading from the hormone to gene expression, it is not sufficient to cause such expression.} \\ \hline
         SEN&\multicolumn{2}{p{0.86\linewidth}}{We conclude that, although the \textcolor{violet}{<el> pH <le>}-induced the \textcolor{blue}{<re> ABA <er>}(i) increase is correlated with and even precedes the induction of RAB-16 mRNA expression and is an essential component of the transduction pathway leading from the hormone to gene expression, it is not sufficient to cause such expression.} \\ \hline
         SEP& \multicolumn{2}{p{0.86\linewidth}}{We conclude that, although the \textcolor{blue}{<re> pH <er>}-induced the \textcolor{violet}{<el> ABA <le>}(i) increase is correlated with and even precedes the induction of RAB-16 mRNA expression and is an essential component of the transduction pathway leading from the hormone to gene expression, it is not sufficient to cause such expression.} \\ \hline
         SREO & \multicolumn{2}{p{0.86\linewidth}}{We conclude that, although the \textcolor{violet}{<el> integrin <le>}-induced the \textcolor{blue}{<re> pH <er>}(i) increase is correlated with and even precedes the induction of RAB-16 mRNA expression and is an essential component of the transduction pathway leading from the hormone to gene expression, it is not sufficient to cause such expression.} \\ \hline
         VNeg &\multicolumn{2}{p{0.86\linewidth}}{We conclude that, although the \textcolor{violet}{<el> ABA <le>}-induced the \textcolor{blue}{<re> pH <er>}(i) increase \textcolor{red}{is not correlated} with and even precedes the induction of RAB-16 mRNA expression and is an essential component of the transduction pathway leading from the hormone to gene expression, it is not sufficient to cause such expression.} \\ \hline
         LPR &\multicolumn{2}{p{0.86\linewidth}}{We conclude that, although the \textcolor{violet}{<el> ABA <le>}-induced the \textcolor{blue}{<re> pH <er>}(i) \textcolor{red}{decrease} is correlated with and even precedes the induction of RAB-16 mRNA expression and is an essential component of the transduction pathway leading from the hormone to gene expression, it is not sufficient to cause such expression.} \\ \hline
         Generation &\multicolumn{2}{p{0.86\linewidth}}{We conclude that the \textcolor{blue}{<re> ABA <er>} -induced increase in \textcolor{violet}{<el> pH <le>} (i) precedes the expression of RAB-16 mRNA.}
    \end{tabular}
    \caption{Example of the generated adversarial instance for the \dataset dataset using lexico semantic constraints. Regulator entities are enclosed in \textcolor{blue}{\textit{<re> <er>}} tags and regulated entities are enclosed in  \textcolor{violet}{\textit{<el> <le>}} tags. The red texts show the negated phrases.} 
    \label{tab:data_example}
\end{table*}
 \section{Evaluation}
 \label{exper}
 
 In this section we benchmark the performance of state-of-the-art (SOTA) biomedical language models on the \dataset dataset. Our evaluation is aimed at assessing the following aspects of the \dataset:
 \begin{enumerate}[noitemsep]
     \item How difficult is the inference task captured by the dataset?
     \item What kinds of perturbations are difficult for the models?
     \item How consistent are the models on adversarial instances?
 \end{enumerate}
 
\subsection{Implementation Details }
We fine-tune two state-of-the-art models in biomedical domain:\\ (i) PubMedBERT~\cite{pubmedbert} was pretrained from scratch with texts from the biomedical domain and shown to be effective for a wide variety of biomedical NLP tasks including NER, QA, and sentence similarity.\\
 (ii) BioLinkBERT~\cite{yasunaga2022linkbert} augments PubMedBERT by pre-training jointly on linked biomedical articles.  \\
 We fine-tune the top 3 layers of the base-sized pretrained models from Hugging Face~\cite{wolf2019huggingface} using PyTorch~\cite{paszke2019pytorch}. We use AdamW~\cite{loshchilov2017decoupled} with a learning rate of $1e-4$ by manually tuning 5 different values. We use the original hyper-parameters of the models and the NLI label prediction is done via binary classification using \texttt{CLS} token. The sequence length we use here is 512 and the beam size is 16. We train each model for 20 epochs and choose the best one based on the performance (macro F1) over the dev set.
 
 
 The performance of both these models is listed in Table~\ref{tab:per_class_perf}.
 The table reports positive, negative, and overall F1 scores, as well performance for the various types of negative examples (recall). At the time of prediction, we use a binary classifier. Hence, we don't have a fine-grained negative category predictions, therefore, we can't calculate precision. Instead, we report recall for fine-grained negative categories and F1 score for positive and overall negative prediction. We also report macro-F1 for positive and negative classes.
 In addition, the table includes an ablation experiment, where the performance of the classifiers trained only on the hypothesis (``hypo-only'') is contrasted with the classifier trained on the entire data (``premise+hyp''). 
 We also include the performance of the balanced distribution in parentheses, to compare with the overall distribution.

\renewcommand{\arraystretch}{1.3}
\setlength\tabcolsep{6.8pt}
 \begin{table*}[ht!]
    \centering
    \begin{tabular}{c|c|c||c|c|c|c||c|c|c|c}
    \multicolumn{3}{c||}{}&\multicolumn{4}{c||}{Full Distribution}  &\multicolumn{4}{c}{Balanced Distribution} \\ \hline
     \multicolumn{3}{c||}{Model}&\multicolumn{2}{c}{PubMedBERT}  &\multicolumn{2}{|c||}{BioLinkBERT}&\multicolumn{2}{c}{PubMedBERT}  &\multicolumn{2}{|c}{BioLinkBERT}\\ \hline
        \multicolumn{3}{c||}{Class}&h-only&p+h&h-only&p+h&h-only&p+h&h-only&p+h \\ \hline \hline
        \multicolumn{3}{c||}{Positive}&0.65 &0.76 &0.68&0.77&0.57&0.69&0.60&0.69\\ \hline
        \multirow{10}{*}{\rotatebox[origin=c]{90}{Negative}} &{\multirow{8}{*}{\rotatebox[origin=c]{90}{Rule-based}}}&SEN&0.90&0.96&0.91&0.97&0.88&0.92&0.93&0.96 \\
        && SEP&0.92&0.97 &0.93&0.98&0.89&0.92&0.93&0.96\\
        && SRE &0.33&0.50&0.33&0.50&0.50&0.67&0.33&0.67\\
        && SREO &0.69 & 0.98&0.64&0.99&0.60&0.95&0.69&0.97\\
        &&VNeg &0.81&0.91&0.82&0.86&0.79&0.84&0.78&0.83\\
        &&SN&0.64&0.82 &0.54&0.81&0.52&0.78&0.59&0.82\\
        &&LPR&0.56&0.56&0.48&0.59&0.50&0.50&0.48&0.49\\\cline{3-11}
        &&{Macro-avg}&0.69&0.81&0.67&0.82&0.67&0.80&0.68&0.81\\\cline{2-11}
        &{\multirow{3}{*}{\rotatebox[origin=c]{90}{NN-based}}}&GEN-ND&0.47&0.6&0.53&0.56&0.57&0.66&0.64&0.68\\
        &&GEN&0.33&0.57&0.33&0.57&0.57&0.68&0.33&0.68\\ \cline{3-11}
        &&{Macro-avg}&0.40&0.58&0.43&0.56&0.57&0.67&0.49&0.68\\\cline{2-11}
        &\multicolumn{2}{c||}{All negatives}&0.63 &0.76 &0.61 & 0.76&0.65&0.77&0.63&0.79\\ \hline \hline
        \multicolumn{3}{c||}{Macro-avg}&0.64&0.76&0.65&0.77&0.61&0.73&0.62&0.74

    \end{tabular}
    \caption{Overall performance of two state-of-the-art models in the biomedical domain (PubMedBERT, BioLinkBERT) on both distributions. The models are fine-tuned using the data with premise (p+h) and without premise (h-only) on the \dataset dataset. The metric used here is recall for fine-grained negative classes and F1 for positive and all negative categories.  The different rows indicate the performance for the various kinds of positive and negative examples. }
    \label{tab:per_class_perf}
\end{table*}

 \subsection{Overall Difficulty}
 Table~\ref{tab:per_class_perf} indicates that the overall performance of the best model on the positive class is 77\%, and 79\% on all negative examples  (macro-average). If we only consider the difficult negative classes (SRE, LPR, GEN-ND, GEN), the best model's performance on the negative categories drop considerably to 55.4\%, i.e.,  only slightly better than a random classifier.

This table also highlights the difficulty of the generated categories. While traditional approaches of the adversarial example creation, (i.e., SEN and SEP), are solvable with large transformer-based models, the more complex negative examples produced using generation are considerably more difficult to be classified correctly.

Table~\ref{tab:per_class_perf} calls attention to another feature of the \dataset dataset: classifiers trained with the balanced training data (shown in parentheticals) perform better on minority categories, while the models produced after training on the larger distribution perform better on other categories. This highlights the versatility of the dataset, as well as the importance of customizing the data distribution (including that of negative examples!) for each use case.

 
 

\subsection{Difficulty of Adversarial Instances}

 Table~\ref{tab:per_class_perf} indicates that some categories of the adversarial examples are more difficult than the rest. This is mostly seen in rule-based categories. In average, the neural-based methods generate more difficult sentences than the rule-based methods.
 
 For instance, SEP and SEN instances are easier due to the structure (markers) in the dataset. Even without inspecting the supporting set, the model learns that the entity with \textit{<re><er>} markers should be the subject of the text while the entity marked with \textit{<el><le>} should be the object of the mechanism. 
 
Some categories are easier to recognize with context. For instance SREO and SN approaches are not easily detectable with hypothesis only baselines. But, when the model is trained with both premise and hypothesis, these become easier because the contradiction can be recognized using information in the premise (i.e. abstract). 

We have four difficult categories of adversarial examples (SRE, LPR, GEN-ND, and GEN) where the models perform only slightly better than a random classifier. 
These are the least-frequent classes; when we train the model under the balanced distribution the performance improves, somewhat. However, performance remains low, which underscores the need for further research on handling these difficult adversarial examples.

\subsection{Model Consistency on Adversarial Instances}
\begin{figure}
     \centering
     \includegraphics[width=0.48\textwidth]{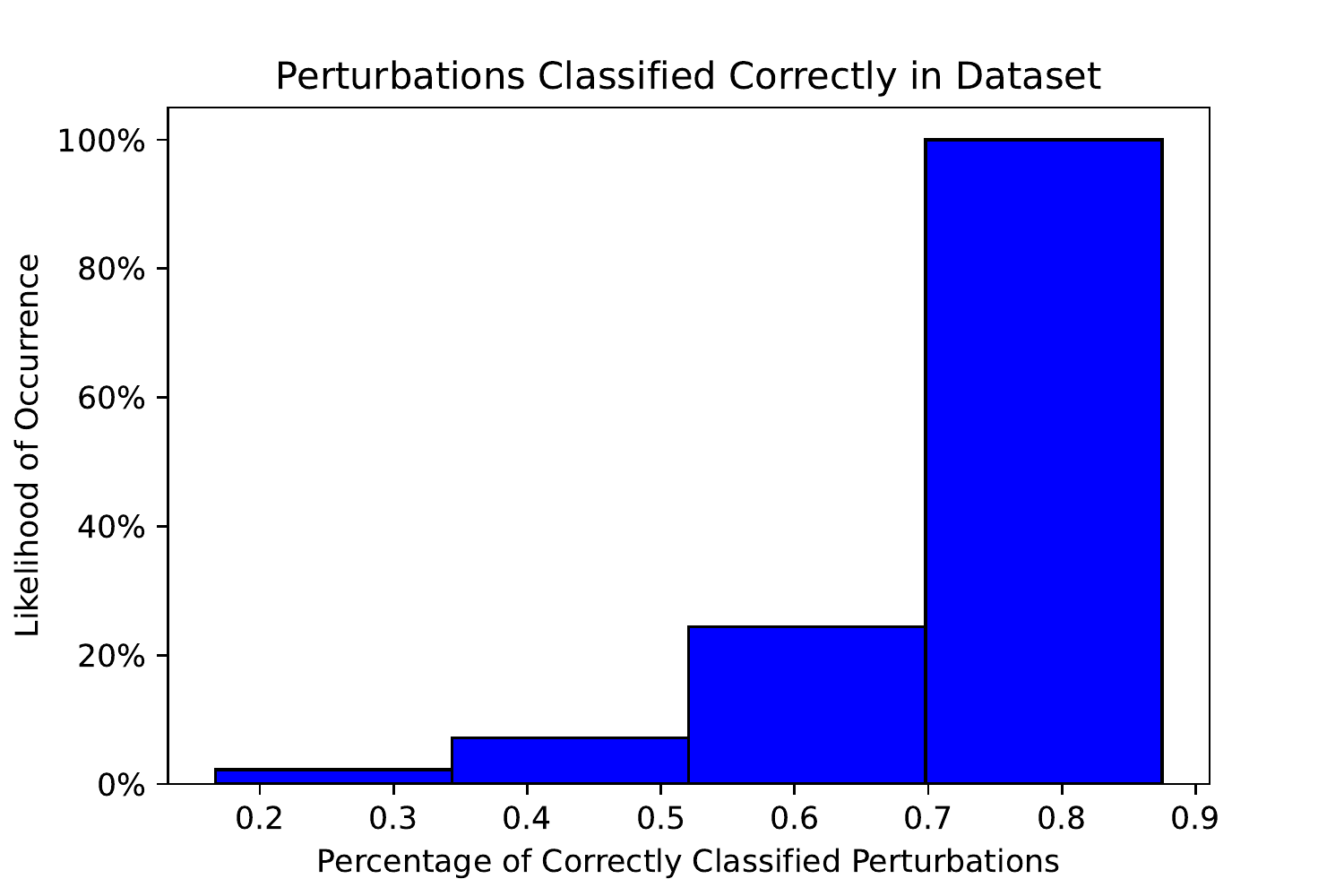}
     \caption{Percentage of correctly classified perturbations in the dev set.}
     \label{fig:class_corr}
 \end{figure}

In addition to the per-perturbation evaluation, we also merged all available positive and negative instance for each entry of the dataset, and computed what percentage of them are classified correctly. The cumulative results are shown in Figure~\ref{fig:class_corr}. While models are able to get reasonable accuracy overall, they are not consistent in their decisions. There are no cases, where both the positive instance and all of its associated negative instances are all classified correctly. There are only $30\%$ of the cases where at least $70\%$ of the perturbations derived from the same positive instance are classified correctly. This further indicates the brittleness (or lack of robust reasoning) in current models suggesting avenues for further research.
 
\renewcommand{\arraystretch}{1}
\begin{table}
    \centering
    \begin{tabular}{c|c}
    Error Category & Frequency $\%$ \\ \hline
        Multiple pieces of information &  $10$ \\
        Abbreviation  & $10$ \\
        Unrelated information & $16$ \\
        Mechanism mix up & $18$ \\
        Noun phrase mix up & $20$ \\
        Entity Similarity & $26$
        
    \end{tabular}
    \caption{Distribution of different error categories in 50 incorrect classified samples}
    \label{tab:error_ana}
\end{table}

\subsection{Error Analysis}
We analyzed a set of 50 instances within the \dataset dataset which are classified incorrectly. \autoref{tab:error_ana} outlines distribution of the errors made by BioLinkBERT model in different categories. In $10\%$ of the misclassified instances, the mechanism behind the entities which is explained in the conclusion sentence needs multi hop reasoning which makes it difficult to classify correctly. 
Abbreviations can cause difficulties, when one of the main entities is mentioned in its full form in one sentence but abbreviated in the rest of the sentences. Distant entities also make inference harder. The model fails when the two main entities are in different sentences with many unrelated fragments of text occur between them. Sometimes the abstracts can talk about two related experiments on the same entities, but the mechanism is only related to one of the experiment making it harder to assess entailment. A '-' (dash) between two words, can change the subject and object of a sentence. The entailment is very difficult when there is a subtle difference here.
 Finally, the similarity between entity names, or name overlaps, specifically in case of SRE and GEN-ND-SRE confuses the model about the correct entailment class. There are cases where the entity is swapped with another entity with similar name or with partially overlapped name, these cases seem to be difficult for the model to classify correctly.

\section{Conclusion}

In this paper, we introduced a novel semi-supervised procedure for the creation of biomedical NLI datasets that include mechanistic information. Our key contribution is automating the creation of negative examples that are informative without being simplistic. We achieve this by manipulating the lexico-semantic constraints in the mechanism structures captured in the hypotheses, which we implement both with rules and with neuro-logic decoding. To our knowledge, this is the first paper that employs neuro-logic decoding for the generation of adversarial examples. All in all, we implemented nine different strategies for the creation of adversarial examples.

We used this procedure to create the \dataset dataset, which addresses NLI for mechanistic texts in the biomedical domain. An evaluation of a sample of these negative examples by human biomedical experts indicated that $86\%$ of these examples are indeed true negatives. We trained two state-of-the-art neural NLI classifiers on this dataset, and showed that the overall performance remains relatively low, which indicates that this NLI task is not solved. Critically, we observe that the performance on the different classes of negative examples varies widely, from $97\%$ accuracy on the simple negative examples that change the role of the entities in the hypothesis, to $55\%$ (i.e., barely better than chance) on the negative examples generated using neuro-logic decoding. We hope that this open-access dataset\footnote{Code and data is available at~\url{https://github.com/StonyBrookNLP/BioNLI}} will enable further research both on biomedical NLI, and on language understanding in general.

\section{Acknowledgement}
This material is based on research that is sup- ported in part by the National Science Foundation under the award IIS \#2007290. The authors would like to thank the anonymous reviewers and the area chair for their feedback on this work. We would also like to thank the biomedical experts who assessed the quality of the adversarial examples. 
\section{Limitations}
Unlike many scientific NLI datasets~\cite{romanov2018lessons,shivade2015textual} no instance in the \dataset dataset was directly annotated by human domain experts. 
Instead, following the trend of machine-generated datasets~\citep{hartvigsen2022toxigen}, we build upon recent developments in text generation and generate \dataset automatically.

The only human annotation in this effort was performed by one expert on a sample of 50 sentences, to check the quality of automatically created negative examples.
This minimal effort was justified by previous work in the biomedical space (citation hidden for review), in which we observed that experts had high inter-annotator agreement on the interpretations of scientific information in abstracts.

The premise-free experiments show the presence of artifacts in some categories of the \dataset dataset, similar to several other NLI datasets~\cite{romanov2018lessons,snli:emnlp2015,nangia2017repeval}. Addressing these artifacts remains an open research issue.

\section{Ethical Considerations}

Our data is collected solely from open-access publications in PubMed. We do not include any meta data (authors, publication venue, etc.) in the dataset. The created dataset is also open-access.


We believe our released dataset and software
will contribute to society by promoting further
NLI research and applications in the biomedical domain. Long term, we envision that this research will enable novel machine reading applications that automatically discover potential explanations and treatments for diseases that are still misunderstood today. 


\bibliography{anthology,custom}
\bibliographystyle{acl_natbib}

\appendix
\label{sec:appendix}

\section{Extraction Patterns for Identifying Conclusion Sentences}
To extract the conclusion sentences from the abstract, we follow the recipe from ~\citet{bastan2022sume}. We filtered the abstracts from PubMed dataset which have a form of conclusion sentence at the end. In particular we filtered out all the abstracts that do not have any of the phrases in Table~\ref{tab:patterns}.

\begin{table}[ht!]
    \centering
    \begin{tabular}{|c|}\hline
        Used Phrase   \\ \hline \hline
        \textit{we conclude that} \\ \hline
         \textit{it is concluded that} \\ \hline
        \textit{it was concluded that}\\ \hline
    \textit{we concluded that} \\ \hline
    \textit{we have concluded that}  \\ \hline
    \textit{it has been concluded that } \\ \hline
  \textit{it may be concluded that }  \\ \hline
   \textit{it was therefore concluded that }  \\ \hline
   \textit{we therefore conclude that }  \\ \hline
\textit{we conclude }  \\ \hline
   \textit{we thus conclude that} \\ \hline
 \textit{it is therefore concluded that} \\ \hline
\textit{we further conclude that} \\ \hline
          
    \end{tabular}
    \caption{Used phrases to filter the abstracts}
    \label{tab:patterns}
\end{table}

\section{Hyper-parameter Selection}
\subsection{Generation Lambda}
One of the strategies to generate negative examples is the  generation method (GEN). We trained a t5-large model on the SuMe dataset using 5-fold cross validation. Each time we trained on 4 folds and generated the output for the 5th fold. From the generated texts, we selected the ones which have lowest quality. That is, we selected generated sentences that contain both entities, the predicted relation labels are incorrect, and the Bleurt score of the generated sentence against the true mechanism sentence is lower than a threshold $\lambda$. In our preliminary experiments we found that $ \lambda=0.45 $ yields a good compromise between quality and yield. Analyzing the output of this hyper-parameter showed that $90\%$ of the sentences selected with this method are indeed true negative samples.

\subsection{Neurological Decoding Hyper-parameters}
One of the strategies to generate negative examples is the  generation method with the neurological decoding (GEN-ND)~\cite{lu-etal-2021-neurologic}. We used the source code introduced in their GitHub page\footnote{\url{http://github.com/GXimingLu/neurologic_decoding}}. The hyper-parameter details are shown in Table~\ref{tab:hyperparamaters}.

We also allowed for the use of negative or positive constraints, another choice that we use as a hyper-parameter.

\renewcommand{\arraystretch}{1.}
\setlength\tabcolsep{4pt}
\begin{table}
    \centering
    \begin{tabular}{c|c|c} 
        Parameter & Value & Details\\ \hline 
        min\_tgt\_length & 15 & min target length \\
        max\_tgt\_length & 256& max target length \\
        bs& 4& batch size \\
        beam\_size &50&beam size \\
        length\_penalty & 0.1 & length penalty for beam \\
        ngram\_size & 10 & ngrams occur once \\
        prune\_factor &50&candidates to keep \\
        sat\_tolerance &2&min satisfied constraints\\
        beta &2& reward factor
    \end{tabular}
    \caption{Neurological decoding hyper parameters}
    \label{tab:hyperparamaters}
\end{table}

\end{document}